\documentclass{article} 
\usepackage{iclr2023_conference,times}

\usepackage{amsmath,amsfonts,bm}

\def\eqref#1{equation~\ref{#1}}

\def\1{\bm{1}}

\DeclareMathAlphabet{\mathsfit}{\encodingdefault}{\sfdefault}{m}{sl}
\SetMathAlphabet{\mathsfit}{bold}{\encodingdefault}{\sfdefault}{bx}{n}

\usepackage{hyperref}
\usepackage{url}
\usepackage{graphicx}
\usepackage{float}
\usepackage{subfig}
\usepackage{array}
\usepackage{siunitx}
\usepackage{color}
\usepackage{colortbl}
\usepackage{graphicx}
\usepackage{enumitem}
\usepackage{tablefootnote}
\usepackage{authblk}

\definecolor{cyan}{cmyk}{.3,0,0,0}
\definecolor{hr}{gray}{0.5}

\def\onedot{.}
\def\eg{{\em e.g}\onedot} 
\def\ie{{\em i.e}\onedot}

\title{Self-supervised Video Representation Learning with Motion-Aware Masked Autoencoders}

\author{
\textbf{Haosen Yang\textsuperscript{\rm 1*} \ Deng Huang \ Bin Wen\textsuperscript{\rm 2} \ Jiannan Wu\textsuperscript{\rm 3}\ Hongxun Yao\textsuperscript{\rm 4} \\ }
\textbf{Yi Jiang\textsuperscript{\rm 2, $\dagger$} \ 
Xiatian Zhu\textsuperscript{\rm 1,  $\dagger$} \ 
Zehuan Yuan\textsuperscript{\rm 2}}

\textsuperscript{\rm 1}University of Surrey \ \textsuperscript{\rm 2}ByteDance \ \textsuperscript{\rm 3}The University of Hong Kong \ \textsuperscript{\rm 4}Harbin Institute of Technology
}

\newlength\savewidth\newcommand\shline{\noalign{\global\savewidth\arrayrulewidth
		\global\arrayrulewidth 1pt}\hline\noalign{\global\arrayrulewidth\savewidth}}

\newcommand{\x}{{\times}}
\def\x{$\times$}

\newcommand{\tablestyle}[2]{\setlength{\tabcolsep}{#1}\renewcommand{\arraystretch}{#2}\centering\footnotesize}
\newcolumntype{*}{>{\global\let\currentrowstyle\relax}}
\newcolumntype{^}{>{\currentrowstyle}}
\newcommand{\rowstyle}[1]{\gdef\currentrowstyle{#1}#1\ignorespaces}

\newcolumntype{x}[1]{>{\centering\arraybackslash}p{#1pt}}
\newcolumntype{y}[1]{>{\raggedright\arraybackslash}p{#1pt}}
\newcolumntype{z}[1]{>{\raggedleft\arraybackslash}p{#1pt}}

\begin{document}
\maketitle

{\let\thefootnote\relax\footnotetext{~$^*$ Haosen Yang was a research intern at ByteDance. Email:\href{haosen.yang.6@gmail.com}{\color{black}{haosen.yang.6@gmail.com}}}}
{\let\thefootnote\relax\footnotetext{ ~$^\dagger$ Corresponding authors: 
\href{mailto:jiangyi.enjoy@bytedance.com}{\color{black}{jiangyi.enjoy@bytedance.com}}, \href{xiatian.zhu@surrey.ac.uk}{\color{black}{xiatian.zhu@surrey.ac.uk}}}}
\begin{abstract}

{Masked autoencoders (MAEs) have emerged recently as art self-supervised spatiotemporal representation learners.}
Inheriting from the image counterparts, however, existing video MAEs 
still focus largely on static appearance learning whilst 
are limited in learning dynamic temporal information hence less effective for video downstream tasks.
To resolve this drawback, 
in this work we present a motion-aware variant -- \texttt{MotionMAE}.
Apart from learning to reconstruct individual masked patches of video frames,
our model is designed to additionally predict the corresponding motion structure information over time. 
This motion information is available at the temporal difference of nearby frames.
As a result, our model can extract effectively both static appearance and dynamic motion spontaneously, leading to superior spatiotemporal representation learning capability.
Extensive experiments show that our \texttt{MotionMAE} outperforms significantly both supervised learning baseline and state-of-the-art MAE alternatives,
under both domain-specific and domain-generic {\em pretraining}-then-{\em finetuning} settings.
In particular, when using ViT-B as the backbone our \texttt{MotionMAE} surpasses the prior art model by a margin of {1.2\%} on Something-Something V2 and {3.2\%} on UCF101 in domain-specific pretraining setting. 
Encouragingly, it also surpasses the competing MAEs
by a large margin of over 3\% on the challenging video object segmentation task.
The code is available at 
\href{https://github.com/happy-hsy/MotionMAE}{\color{blue}{https://github.com/happy-hsy/MotionMAE}}
\end{abstract}
\section{Introduction}


Masked (denoising) autoencoding (MAE) \citep{vincent2010stacked,vincent2008extracting} has resurged as the state-of-the-art self-supervised image representation learning approach \citep{he2022masked,atito2021sit,bao2021beit,wei2022masked,dosovitskiy2020image,li2021mst}.
This is inspired by the remarkable success of BERT \citep{devlin2018bert} in natural language processing. 
The idea of MAE is masked token/patch recovery within the Transformer framework \citep{vaswani2017attention}.
Existing image MAE models focus on visual tokenization \citet{bao2021beit,dong2021peco},
token masking strategy \citep{li2021mst,atito2021sit},
reconstruction target \citep{wei2022masked},
and architectural efficiency \citep{he2022masked}.
Recently, MAE has been also applied to learn more challenging spatiotemporal representations from videos \citep{feichtenhofer2022masked,tong2022videomae,wang2022bevt}.
Despite demonstrating strong downstream task performances, we conjugate that these methods are still suboptimal in learning temporal structure information, since their learning objective is largely limited to the reconstruction of {\em individual tokens of video frames}, resulting in an over-challenging task of learning dynamic temporal information across frames.

\begin{figure}[t]
\vspace{-2em}
\centering
\includegraphics[width=0.95\linewidth]{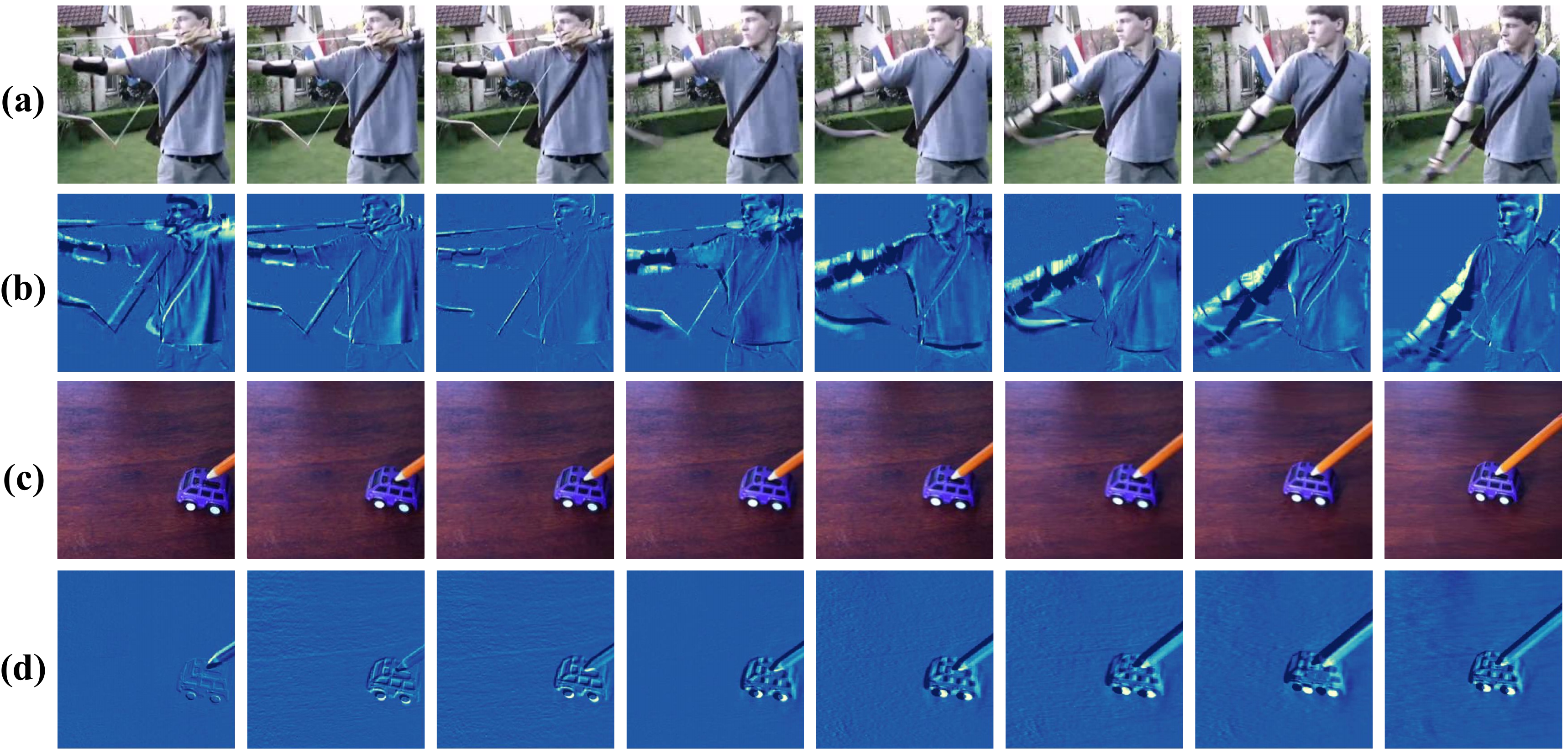}
\caption{\textbf{Illustration of local motion information in video data}. Whilst {\bf (a, c)} every single video frame  provides rich appearance information, 
{\bf (b, d)} their temporal difference gives additional dynamic motion knowledge over time. Intuitively, appearance and motion both are critical in learning spatiotemporal representation for video understanding. 
}
\label{fig:tease}
\end{figure}

To overcome the aforementioned limitation, in this work a simple yet superior video self-supervised learning method, dubbed as {\bf\em MotionMAE}, is introduced. Our idea is {\em local motion reconstruction} -- recovering the motion information corresponding to each masked frame patch.
The target motion information is obtained by simply contrasting two temporally adjacent video frames (\ie, the temporal difference of video frames). 
This is an intrinsic temporal ingredient of video data, in addition to the spatial appearance of frame images.
As illustrated in Fig. \ref{fig:tease}, 
the frame difference presents detailed motion information of
push-up over time.
Specifically, given a training video clip, \texttt{MotionMAE} randomly masks out a certain percentage (\eg, 90\%) of patches per frame, and reconstructs both masked frame patches for spacial appearance modeling and the corresponding motion structure for temporal dynamics modeling concurrently. 
For computational efficiency, we adopt an asymmetric MAE architecture \citep{he2022masked} based on vanilla Vision Transformers (ViTs) \citep{dosovitskiy2020image}: The encoder operates only on visible patches and the decoder on all the patches.

{In a nutshell, we propose a motion-aware MAE method, \texttt{MotionMAE}, for self-supervised spatiotemporal representation learning from unlabeled videos.}
For extensive evaluation, we conduct both domain-generic and domain-specific {\em pretraining}-then-{\em finetuning} paradigms.
The results show that our \texttt{MotionMAE} outperforms both supervised learning and alternative state-of-the-art MAE designs on action classification, often by a large margin.
In particular, \texttt{MotionMAE} achieves an accuracy of 75.5\% on Something-Something V2, 85.3\% on Kinetics-400, 
and 94.0\% on UCF101 under domain-specific pretraining setting. 
The performance advantage of our model 
remains on the challenging Video Object Segmentation task with a margin of 3+\% over the competing MAE models on the DAVIS2017 benchmark.

\section{Related Work}

\paragraph{Self-supervised learning of video representation}
Self-supervised learning (SSL) of visual representation \citep{he2020momentum, grill2020bootstrap, chen2020simple, caron2020unsupervised} have advanced massive recently and demonstrated superior performance over previously dominant supervised learning (\eg, on ImageNet with exhaustive manual labeling). 
This opens up a door to achieve ever stronger representations
due to the availability of much larger volume of  (even infinite) unlabeled visual data from diverse sources.
This inspires SSL for video data with extra temporal dimension
as compared to static images.
In the literature, the previous waive of video SSL methods rely on carefully designing time-related pretext tasks in three lines:
(1) Predicting specific transformations (\emph{e.g.}, rotation angle \citep{jing2018self}, playback speed \citep{benaim2020speednet}, temporal order \citep{misra2016shuffle, lee2017unsupervised,xu2019self} and  motion statistics \citep{wang2019self}); 
(2) Predicting future frame \citep{han2019video,han2020memory};
(3) Instance discrimination \citep{qian2021spatiotemporal,wang2020self,chen2021rspnet}.
Usually, top-performing methods need to conduct strong and diverse data augmentations, which brings about the notorious scalability bottleneck and learning complexity.
Also, the model performance is conditioned on a good number of 
tricks (\eg, feature bank, data augmentation design).
\paragraph{Masked autoencoders}
With the increasingly wide adoption of Transformers \citep{vaswani2017attention} in computer vision, masked autoencoders (MAEs) have recently emerged as a general SSL framework \citep{he2022masked,bao2021beit,radford2018improving}.
This is mostly inspired by the success of BERT~\citep{devlin2018bert}
in a wide range of NLP downstream tasks.
In general, it can be also regarded as a special pretext task.
However, compared to most previous alternatives, 
MAEs are not only simpler in design,
but also stronger in representation learning.
Specifically, image MAE variants learn the representation
by predicting the masked/unknown regions from visible parts.
For example, iGPT~\citep{chen2020generative} operates on sequences of pixels and predicts unknown pixels.
ViT~\citep{dosovitskiy2020image} predicts the mean colors of masked patches.
BEiT~\citep{bao2021beit} further improves MAEs performance with masked visual token prediction. 
PeCo~\citep{dong2021peco} suggests to inject perceptual similarity during visual codebook learning. 
Interestingly, MAE~\citep{he2022masked} demonstrates the strength of the straightforward idea of image patch reconstruction, in addition to improving the pretraining efficiency by adopting high masking ratios and encoding only unmasked patches.
Alternatively, MaskFeat~\citep{wei2022masked} leverages HOG~\citep{dalal2005histograms} feature as the prediction target to yield strong visual representation.

Very recently, MAE has been similarly applied for learning more challenging spatiotemporal representations from videos.
For example, by extending the BEiT pretraining paradigm,
BEVT~\citep{wang2022bevt} decouples masked SSL into spatial representation learning on images and temporal dynamics learning
on both images and videos in a two-stage process.
%
VideoMAE~\citep{tong2022videomae} and MAE~\citep{feichtenhofer2022masked} 
simply reconstruct masked spatiotemporal patches of each video
against even high masking ratios (\eg, 90\%)
and achieve strong performance on video downstream tasks.
This result is encouraging and inspiring given their simplicity.
However, we consider that such straight inheriting from image MAEs 
would be limited in learning temporal structure information.
This is because their learning objective tends to focus on reconstruction of {\em individual tokens of video frames}, since leaving the model itself to derive the underlying implicit spatiotemporal representation is drastically challenging.
To tackle this problem, we present a motion-aware MAE
that additionally predicts the corresponding motion structure information.
The target motion to be reconstructed is obtained by contrasting two temporally adjacent video frames (\ie, the temporal difference of video frames). 
Extensive experiments and analysis validate our proposed method.
\section{Method}

\begin{figure}[t]
\centering
\vspace{-3em}
\includegraphics[width=0.95\linewidth]{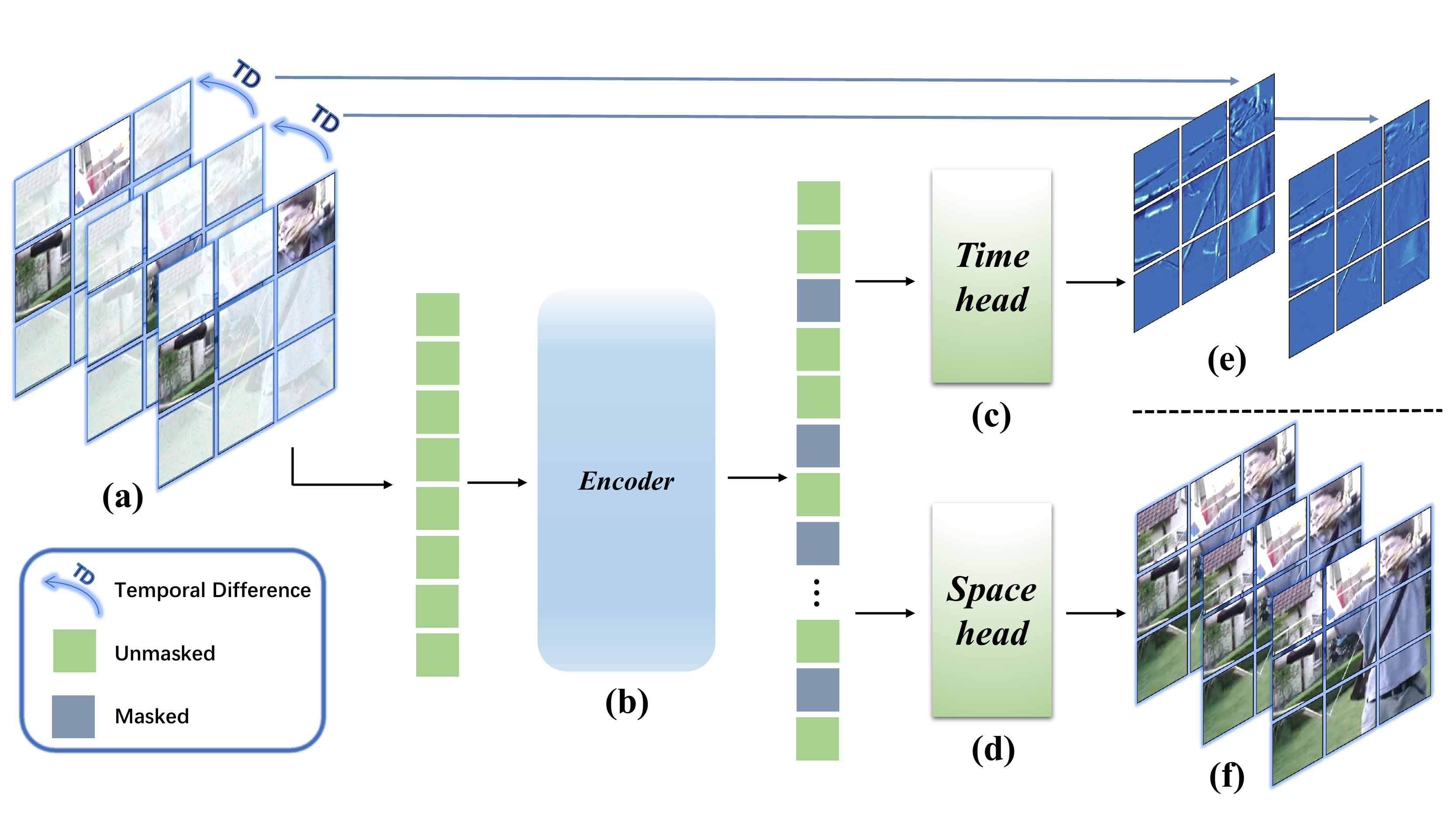}
\caption{ 
Illustration of the proposed  \texttt{MotionMAE.} 
{\bf(a)} We randomly mask the patches of each frame of a given video clip at a fixed ration.  
In {\em pretraining}, {\bf(b)} an encoder operates on the set of visible patches. {\bf(c)} A time head and {\bf(d)} a space head then processes the full set of encoded patches and mask tokens to reconstruct {\bf(e)} the masked frame patches and {\bf(f)} the corresponding local motion, concurrently.
After pretraining, we discard the time head and space head then evaluate the pretraining quality of the encoder on downstream task. 
}
\label{fig:framework}
\end{figure}

Our proposed \texttt{MotionMAE} adopts the asymmetric Transformer based masked autoencoder architecture \citep{he2022masked}, 
as depicted in Fig. \ref{fig:framework}.
The objective is to {\em pretrain a video representation model} in a self-supervised manner
for facilitating downstream tasks such as action classification.
Next, we describe the details of our model architecture and pretraining.

\paragraph{Patch embedding} 
We divide evenly an input video into a regular grid of non-overlapping spacetime patches (\ie, cubes)
and perform 
cube embedding \citep{bertasius2021space,arnab2021vivit}. 
This reduces the temporal resolution of video to suppress
the redundancy degree over time.
Learnable positional embeddings are applied via elementwise addition.

\paragraph{Patch masking}
We randomly mask the patches at each time point of a video clip at a fixed ratio.
Common masking strategies include random, space-only (\ie, tube), and time-only (masking the whole frames selected).
As video data present more redundancy than both text and image, setting a higher masking ratio (\eg, 90\%) is helpful.

\paragraph{Autoencoding}
We adopt a vanilla ViT \citep{dosovitskiy2020image} as the encoder with space-time joint attention. 
Following \citep{he2022masked,feichtenhofer2022masked}, the encoder is applied only on unmasked patches.
This greatly reduces the time and memory complexity, leading to a more efficient solution. 
The encoder can capture high-level spatiotemporal semantic information by interacting patches of all frames.

Our decoder is another vanilla ViT deployed on the union of the encoded patch set and a set of mask tokens \citep{he2022masked}.
It consists of two heads in parallel: 
{\bf\em a space head} to predict masked frame patches,
and {\bf\em a time head} to predict the corresponding local motion structure.
The motion target is obtained by computing $L_1$ difference of pixel values between two temporally nearby frames, which encodes the short-term temporal dynamics information underlying across frames.
This dynamics information underlies along the time dimension
which is complementary to the appearance information of individual video frames.
Our motivation is that by decomposing the two types of information underlying the video data, self-supervised learning of spatiotemporal representation could be facilitated. 
This could be considered as introduction of prior knowledge 
which otherwise needs to be discovered from the learning process itself.
For each decoder head, specific positional embeddings are applied following MAE \citep{he2022masked}. 
Since video data is high-dimensional, the decoder is designed to be more lightweight than the encoder, consuming less computational cost despite processing the full patch set with two heads.
\paragraph{Objective loss function}
For model pretraining, for each head we minimize the $L_2$ reconstruction loss between the prediction and the target (\ie, the frame image patches for the space head and frame difference patches for the time head)
over the masked patches.
The final objective is to minimize their summation.

\section{Implementation}
\label{sec:impl}

\paragraph{Downstream tasks}
We focus on two significant downstream tasks:
action recognition, and video object segmentation.
For action recognition, we evaluate three datasets:
UCF101 \citep{soomro2012ucf101}, Kinetics-400 \citep{kay2017kinetics}, and Something-Something V2 \citep{goyal2017something}. 
Specifically, {\em Kinetics-400} contains around 240k training videos and 20k validation videos at the length of 10 seconds from 400 action classes.
{\em Something-Something V2} offers around 169k videos for training and 20k videos for
validation.  
In contrast to Kinetics-400, this dataset contains
174 motion-centric action classes.
These two large-scale video datasets focus on different visual cues for action recognition.
{\em UCF101} is a small video dataset with around 9.5k/3.5k train/val videos. This allows for testing with limited training data.
All the three datasets together provide a cohort
with varying-sized benchmarks for extensive evaluation.

For video object segmentation, we test the {\em DAVIS-2017} \citep{pont20172017} dataset with 30 videos for validation and 59 object classes.
The task is to segment the salient moving objects in the scene.

\paragraph{Data pre-processing}
For \texttt{MotionMAE} pretraining, by default the input is a clip of 16 frames at a resolution of 224$\times$224 pixels (except otherwise stated). 
A training clip is sampled from a raw video.
We set the temporal stride of 2, 4, and 4 for Something-Something V2, Kinetics-400, and UCF101, respectively, with the starting frame randomly chosen.
For the spatial dimensions, we perform random resized cropping at a scale range of [0:5; 1].
During both pretraining and finetuning, we apply flip augmentation on Kinetics-400 and UCF101, but not on Something-Something V2 as this dataset involves order-sensitive action classes. 
We do not apply any other data augmentations.

\paragraph{Architecture}
We adopt the vanilla ViT architecture \citep{dosovitskiy2020image} as the encoder and decoder, namely the ViT-B and ViT-L variants.
For a patch, we use a temporal size of 2 and a spatial size of 16 \x 16, denoted as 2 \x 16 \x 16.
Given a 16 \x 224 \x 224 input, the corresponding patch size is 8 \x 14 \x 14 tokens.
We do not use the \texttt{[CLS]} token in the ViT models, yielding a small improvement in running-time without performance loss.
We use a Transformer decoder consisting of 4 layers  with 384 and 512 embedding dimensions for ViT-B and ViT-L, respectively. 
The decoder outputs both frame patches and motion patches at two heads.
We use sinusoidal positional encoding in both the encoder and the decoder.

\paragraph{Masking}
During pretraining,
similar as prior works~\citep{tong2022videomae, feichtenhofer2022masked},
we apply a masking ratio of 90\% for large datasets (\eg, Kinetics-400 and Something-Something V2)
and 75\% for small ones (\eg, UCF101).
We find the masking ratio is partly conditioned on the dataset scale.
For the masking strategy, we find the best option for UCF101, Something-Something V2 and Kinetics400
to be random, random and space-only (\ie, tube), respectively.
In general, space-only and random can yield similar performance, whist time-only performs the worst.
We ablate the masking hyper-parameters and strategies in the \texttt{appendix}.

\paragraph{Setting}
For fair comparison, we follow the pretraining configuration of~\citep{tong2022videomae}.
Specifically, we use the AdamW optimizer with a batch size of 1024. 
We evaluate the pretraining quality by {\em end-to-end fine-tuning}. 
In model inference, we adopt the common multi-view practice: 
(1) Taking $K$ temporal clips ($K=2, 5,10$ for Something-Something V2, Kinetics-400 and UCF101) to cover the  whole video length, 
and (2) for each clip, taking 3 spatial views to cover the longer spatial axis,
resulting in a total of $3K$ views. The final prediction is obtained by averaging all the views. 
More implementation details and hyper-parameters are given in the \texttt{appendix}.

\section{Experiments}
\subsection{Action Recognition}
We contrast the proposed \texttt{MotionMAE} with two categories of 
state-of-the-art video representation learning methods:
(1) {\bf Supervised video learning methods} including SlowFast~\citep{feichtenhofer2019slowfast}, TDN~\citep{wang2021tdn}, Timesformer~\citep{bertasius2021space}, ViViT~\citep{arnab2021vivit}, MViT~\citep{fan2021multiscale,li2022mvitv2}, Video Swin~\citep{liu2022video} and MotionFormer~\citep{patrick2021keeping}.
(2) {\bf Self-supervised video learning methods} including 
OPN~\citep{lee2017unsupervised}, VCOP~\citep{xu2019self}, CoCLR~\citep{han2020self}, SpeedNet~\citep{benaim2020speednet}, Pace~\citep{wang2020self}, RSPNet~\citep{chen2021rspnet}, ASCNet~\citep{huang2021ascnet}, MMV~\citep{alayrac2020self}, XDC~\citep{alwassel2020self}, GDT~\citep{patrick2020multi},
BEVT~\citep{wang2022bevt}, Maskfeat~\citep{wei2022masked},
VideoMAE~\citep{tong2022videomae} and MAE~\citep{feichtenhofer2022masked}.

For more extensive comparison, we consider both domain-generic (\ie, training and test on different datasets)
and domain-specific (\ie, training and test on the same dataset) 
pretraining-then-finetuning settings.

\paragraph{Something-Something-V2}
\begin{table}[t]
\centering
\begin{tabular}{*l|^l|^l|^c|^c|^c|^c}
Method &Extra data &Backbone & Frames & Top-1 (\%) & GFLOPs & Param(M) \\
\shline \hline
\rowstyle{\color{hr}} 
SlowFast & K400 & SlowFast & {8+32} & 63.1 & {106}\x{1}\x{3} & 63.1 \\
\rowstyle{\color{hr}} TDN & IN1K & ResNet101 & {8+16} & 69.6 & {198}\x{3}\x{1} & 88 \\
\rowstyle{\color{hr}} TimeSformer & K400+IN21K & ViT-L & {64} & {62.4} & {5549}\x{1}\x{3} & 430 \\
\rowstyle{\color{hr}} ViViT & K400+IN21K & ViT-L & {32} & {65.9} & {995}\x{4}\x{3} & N/A \\
\rowstyle{\color{hr}} MViTv1  & K400 & MViT-B & {64} & {67.7} & {455}\x{3}\x{1} & 37 \\
\rowstyle{\color{hr}} MotionFormer & K400+IN21K & ViT-L & {32} & {68.1} & {1185}\x{1}\x{3} & 382 \\
\rowstyle{\color{hr}}Video Swin & K400+IN21K & Swin-B & {32} & {69.6} & {321}\x{1}\x{3} & 88 \\
\rowstyle{\color{hr}} MViTv2 & K400 + IN21K & MViTv2-L & {40} & 73.3  & {2828}\x{1}\x{3}  & 213\\ 

BEVT & K400 + IN1K & Swin-B &  {32}  & 71.4 & {321}\x{1}\x{3} & 88 \\
MaskFeat & K400 & MViTv2-L & {40} & 74.4  & {2828}\x{1}\x{3} & 218 \\
MAE & K400 & ViT-L & {16} & 72.1 & {598}\x{1}\x{3}  & 304 \\
MAE & K400 & ViT-H & {16} & 74.1  & {1193}\x{1}\x{3} & 632 \\
VideoMAE & - & ViT-B & {16} & 70.6 & {180}\x{2}\x{3}  & 87 \\
VideoMAE & - & ViT-L & {16} & 74.2  & {598}\x{2}\x{3}  & 305 \\
VideoMAE & - & ViT-L & {32} & 75.3 & {1436}\x{2}\x{3}  & 305 \\
\rowcolor{cyan!50}
\textbf{MotionMAE} & K400 & ViT-L & {16} & \textbf{74.3} & {598}\x{2}\x{3}  & 305  \\
\rowcolor{cyan!50}
\textbf{MotionMAE} & - & ViT-B & {16} & \textbf{71.8} & {180}\x{2}\x{3}  & 87  \\
\rowcolor{cyan!50}
\textbf{MotionMAE} & -  & ViT-L & {16} & \textbf{74.6}  & {598}\x{2}\x{3}  & 305 \\
\rowcolor{cyan!50}
\textbf{MotionMAE} & - & ViT-L & {32} & \textbf{75.5} & {598}\x{2}\x{3}  & 305 \\

\end{tabular}
\caption{Comparison with the state-of-the-art methods on {\bf Something-Something V2}. 
Due to varying dataset sizes, we pretrain our \texttt{MotionMAE} by 1600 epochs on Kinetics-400 (K400) and by 2400 epochs on Something-Something V2.
\texttt{IN21K}: ImageNet-21K; \texttt{IN1K}: ImageNet-1K.
GFLOPs format: A single view's $\times$ temporal views $\times$ spatial views. \texttt{N/A}: Not Available. 
\textcolor{gray}{Grayed}: Supervised learning methods.}
\label{tab:ssv2}
\end{table}

On this temporal structure sensitive dataset,
from Table~\ref{tab:ssv2} we draw several observations. 
{\bf(1)} In terms of learning strategy,
self-supervised learning is generally superior 
over the conventional supervised-learning.
For example, MViTV2 \citep{li2022mvitv2} is clearly outnumbered by MaskFeat \citep{wei2022masked} despite using extra data (IN21K).
{\bf{(2)}} Domain-specific pretraining is often more performing than the domain-generic counterpart as reflected in the contrast of VideoMAE \citep{tong2022videomae} and MAE \citep{feichtenhofer2022masked}.
{\bf{(3)}} In either setting, our \texttt{MotionMAE} exceeds all the alternative solutions by a descent margin. 
Specifically, our method outperforms VideoMAE by a range of $[0.2\%, 1.2\%]$ in the domain-specific setting
and MAE by 2.2\% in domain-generic setting.
Besides, ViT-L based \texttt{MotionMAE} achieves the unprecedented top-1 accuracy of \textbf{75.5 \%}.
This validates the advantage of our pretraining method
in learning spatiotemporal representation from unlabeled video data.
\paragraph{Kinetics-400}
\begin{table}[t]
\centering
\begin{tabular}{*l|^l|^l|^c|^c|^c|^c}
Method &Extra data &Backbone & Frames & Top-1 (\%) & GFLOPs & Param(M) \\
\shline \hline
\rowstyle{\color{hr}} SlowFast &IN1K & SlowFast & {16+64} & 79.8 & {234}\x{10}\x{3} & 60 \\
\rowstyle{\color{hr}} TDN  & IN1K & ResNet101 & {8+16} & 79.4 & {198}\x{10}\x{3} & 88 \\
\rowstyle{\color{hr}}TimeSformer & IN21K & ViT-L & {96} & {80.7} & {8353}\x{1}\x{3} & 430 \\
\rowstyle{\color{hr}}ViViT& IN21K & ViT-L & {128} & {81.7} & {3980}\x{3}\x{1} & N/A \\
\rowstyle{\color{hr}}MViT& - & MViT-B & {64} & {81.2} & {455}\x{3}\x{3} & 37 \\
\rowstyle{\color{hr}}MotionFormer & IN21K & ViT-L & {32} & {80.2} & {1185}\x{10}\x{3} & 382 \\
\rowstyle{\color{hr}}Video Swin & IN21K & Swin-L & {32} & {83.1} & {604}\x{1}\x{3} & 197 \\
\rowstyle{\color{hr}}MViTv2 & IN21K & MViTv2-B& {32} & 82.9  & {255}\x{1}\x{3}  & 213\\ 

BEVT & IN1K & Swin-B &  {32}  & 81.1 & {282}\x{4}\x{3} & 88 \\
MaskFeat & -  & MViTv2-L & {40} & 84.3  & {377}\x{10}\x{1} & 218 \\
VideoMAE & - & ViT-B & {16} & 81.5 & {180}\x{5}\x{3}  & 87 \\
VideoMAE& - & ViT-L & {16} & 85.1  & {598}\x{5}\x{3}  & 305 \\
MAE & - & ViT-L & {16} & 84.8 & {598}\x{7}\x{3}  & 304 \\
MAE & - & ViT-H & {16} & 85.1  & {1193}\x{7}\x{3} & 632 \\

\rowcolor{cyan!50}
\textbf{MotionMAE} & - & ViT-B & {16} & \textbf{81.7}  & {180}\x{5}\x{3}  & 87 \\
\rowcolor{cyan!50}
\textbf{MotionMAE} & - & ViT-L & {16} & \textbf{85.3} & {598}\x{5}\x{3}  & 305  \\

\end{tabular}
\caption{Comparison with the state-of-the-art methods on {\bf Kinetics-400} (K400). 
We pretrain our \texttt{MotionMAE} by 1600 epochs on Kinetics-400 (K400).
\texttt{IN21K}: ImageNet-21K; \texttt{IN1K}: ImageNet-1K.
GFLOPs format: A single view's $\times$ temporal views $\times$ spatial views. \texttt{N/A}: Not Available. 
\textcolor{gray}{Grayed}: Supervised learning methods.
}
\label{tab:k400}
\end{table}

Unlike Something-Something V2, appearance information
plays a more dominant role in Kinetics-400.
However, from Table ~\ref{tab:k400} we can still see similar observations.
For example, 
{\bf{(1)}} \texttt{MotionMAE} remains the best pretraining method among all the competitors.
This suggests that dedicated motion recovery from the masking as we introduce is also beneficial.
{\bf{(2)}} By achieving the top-1 accuracy of \textbf{85.3\%}, ViT-L based \texttt{MotionMAE} even surpasses ViT-H based MAE (85.1\%), while enjoying high efficiency in GFLOPs.
These results indicate the generic performance superiority of our method over diverse action genres.  
\paragraph{UCF101}
\begin{table}[t]
\centering
\begin{tabular}{*l|^l|^l|^c|^c|^c|^c}
Method &Extra data &Architecture & Frames & Top-1 (\%)  & Modality & Param(M) \\
\shline \hline

OPN & - & VGG & N/A &59.6 & V & N/A \\
VCOP & - & R(2+1)D & N/A & 72.4 & V & N/A \\
CoCLR & - & S3D-G & {32} & 81.4  & V & 9 \\
SpeedNet & K400 & S3D-G & {64} & 81.1 & V & 9 \\
Pace & K400 & R(2+1)D & {16} & 77.1 & V & 16 \\
RSPNet & K400 & S3D-G & {64} & 93.7 & V  & 9M \\
ASCNet & K400 & S3D-G & {64} & 90.8 & V &  9M\\
\rowstyle{\color{hr}} MMV & AS+HTM & S3D-G & {32} & 92.5 & V+A+T & 9 \\
\rowstyle{\color{hr}} XDC & IG65M & R(2+1)D& {32} & 94.2 & V+A & 15 \\
\rowstyle{\color{hr}} GDT & IG65M & R(2+1)D& {32} & 95.2 & V+A & 15 \\
VideoMAE & - & ViT-B & {16} & 90.8 & V & 305 \\
VideoMAE & K400 & ViT-B & {16} & 96.1 & V&305 \\
\rowcolor{cyan!50}
\textbf{MotionMAE} & - & ViT-B & {16} & \textbf{94.0} & V  & 87 
\\
\rowcolor{cyan!50}
\textbf{MotionMAE} & K400 & ViT-B & {16} & \textbf{96.3} &V & 305 \\

\end{tabular}
\caption{Comparison with the state-of-the-art methods on {\bf UCF101}. 
Due to varying dataset sizes, we pretrain \texttt{MotionMAE} by 2400 epochs on UCF101 and by 1600 epochs on Kinetics-400 (K400).
\texttt{AS}: Audio-Set \citep{gemmeke2017audio}; \texttt{HTM}: HowTo100M \citep{miech2019howto100m}; \texttt{IG65M}: Instagram-65M \citep{ghadiyaram2019large}.
\texttt{V}: Visual; 
\texttt{A}: Audio;
\texttt{T}: Text or narration;
\texttt{N/A}: Not Available.
\textcolor{gray}{Grayed}: Multimodal methods.
}
\label{tab:ucf101}
\end{table}

Following the two large action datasets as above,
we further evaluate on the small UCF101 dataset.
As dataset size is a critical dimension in self-supervised learning.
As shown in Table ~\ref{tab:ucf101}, it is encouraging that our \texttt{MotionMAE} can surpass all the alternative methods in both domain-specific and domain-generic settings. 
For instance, when pretrained on Kinetics-400 (K400),
our method reaches the best ever classification accuracy of \textbf{96.3\%}, higher than the prior art self-suerpvised learning method VideoMAE \citep{tong2022videomae} and multimodal learning method
GDT \citep{patrick2020multi} (despite using much more videos from IG65M, more video frames, and extra audio modality).
This highlights the crucial significance of motion information which, once learned properly as in our proposed pretraining method, would demonstrate stronger representational power than other techniques (\eg, multimodal alignment).
Another highlight is that in the domain-specific setting
characterized by much lower training cost in this congtext, our \texttt{MotionMAE} is favored over the most similar competitor VideoMAE by as large as \textbf{3.2\%}.
These observations further verify the rich advantages of
our method over existing alternatives.

\subsection{Video Object Segmentation}
\begin{table}[t]
\centering
\begin{tabular}{l|c|c|c|c|c}
Method &Pretrain data &Architecture & J\&F-Mean  & J-Mean & F-Mean \\
\shline \hline
MAE & IN1K& ViT-B& 50.9&49.3&52.6 \\
MAE & K400& ViT-B& 53.5&52.6&54.4 \\
VideoMAE & K400 & ViT-B & 53.8 & 53.2 & 54.4 \\ 
\rowcolor{cyan!50}
\textbf{MotionMAE} & K400 & ViT-B & \textbf{56.8} & \textbf{55.8} & \textbf{57.8} \\
\end{tabular}
\caption{Video object segmentation results on DAVIS 2017. 
The pretrained models are frozen {\em without} task-specific finetuning.
\texttt{Metrics}: mean region similarity J-Mean and mean contour-based accuracy F-Mean.
\texttt{IN1K}: ImageNet-1K; 
\texttt{K400}: Kinetics-400.
}
\label{tab:vos}
\end{table}

In addition to action recognition evaluation,
next we evaluate the usefulness of our method in a task of segmenting salient objects in videos.
This presents a different testing perspective 
as it is highly object-centered.
To that end, we compare \texttt{MotionMAE} with two prior art self-supervised video learners (MAE \citep{feichtenhofer2022masked} and VideoMAE \citep{tong2022videomae}).
We freeze the pretrained models to purely evaluate their expressiveness of video objects over space and time, {\em without} any finetuning.

We adopt the same experimental protocol as in~\citep{jabri2020space} by segmenting the scenes with a nearest-neighbor between consecutive frames.
We draw a couple of observations from Table ~\ref{tab:vos}:
\textbf{(1)}
Pretraining data makes difference as exampled by MAE, and video data is generally more effective as expected.
This suggests the importance of learning spatiotemporal representation to this task.
\textbf{(2)}
When the same video dataset is used, our method surpasses both competing MAEs by a margin of 3.0\% or more.
This again validates the superiority of our video representation learning method in the object-focused scenario, thanks to the proposed masked motion information modeling in addition to static appearance information reconstruction.

\subsection{Ablation Study}
\begin{table}[t]
\centering
\subfloat[Reconstruction target.]
{
\begin{minipage}{0.3\linewidth}{
\begin{center}
\begin{tabular}{c | c}
Target &  top-1 (\%)  \\
\shline
Frame  & 67.8   \\
Motion & 67.6 \\
Frame+Motion & \textbf{68.4}\\
\end{tabular}
\label{tab:targets}
\end{center}}
\end{minipage}
}
\subfloat[Weight ratio: frame to motion.]
{
\begin{minipage}{0.4\linewidth}{\begin{center}
\begin{tabular}{c| c }
Ratio &  top-1 (\%) \\
\shline
1:2 & 68.2  \\
1:1 & \textbf{68.4} \\
2:1 & 68.3  \\\end{tabular}
\label{tab:balance}
\end{center}}\end{minipage}
}
\subfloat[Motion granularity (MG).]
{
\begin{minipage}{0.3\linewidth}{\begin{center}
\begin{tabular}{c| c }
MG &  top-1 (\%)  \\
\shline
1 & \textbf{68.4} \\
2 & 67.8 \\
3 &  67.4 \\\end{tabular}
\label{tab:motion range}
\end{center}}\end{minipage}
}
\\
\subfloat[Decoder width.]
{
\begin{minipage}{0.3\linewidth}{
\begin{center}
\begin{tabular}{c| c}
Decoder width &  top-1 (\%)  \\
\shline
128  & 67.5  \\
384  & \textbf{68.4} \\
512  & 68.3 \\
\end{tabular}
\label{tab:dim}
\end{center}
}\end{minipage}
}
\subfloat[Decoder depth.]
{
\begin{minipage}{0.4\linewidth}{
\begin{center}
\begin{tabular}{c | c}
Decoder depth &  top-1 (\%) \\
\shline
1  & 67.4   \\
4  & \textbf{68.4}\\
8  & 68.2\\\end{tabular}
\label{tab:blocks}
\end{center}
}\end{minipage}
}
\subfloat[Reconstruction loss.]
{
\begin{minipage}{0.3\linewidth}{\begin{center}
\begin{tabular}{c | c}
Loss function & top-1 (\%)  \\
\shline
$L_1$  & 67.6  \\
MSE & \textbf{68.4}   \\
Smooth $L_1$ & 67.5  \\
\end{tabular}
\label{tab:loss}
\end{center}}\end{minipage}
}
\caption{Ablation experiments on \textbf{Something-Something V2}
under the domain-specific pretraining setting. 
Pretraining length of our \texttt{MotionMAE}: 400 epochs. 
\texttt{Model}: ViT-B with the input video clip at a resolution of 16$\times$224$\times$224. 
}
\label{tab:ablations}
\end{table}

We perform ablation study on Something-Something V2
in the most-performing domain-specific pretraining setting.
We use ViT-B as the backbone under the inference setting of 2 clips $\times$ 3 crops.

\paragraph{Reconstruction target}
In this study, we conjugate that a key ingredient with MAE is reconstruction target.
To evaluate this, we swap the reconstruction target 
among the choices of $\{\texttt{frame, motion (temporal difference), frame+motion})\}$, whilst keeping all the others.
We summarize a few key observations:
{\bf (1)} As shown in Table~\ref{tab:targets}, the \texttt{frame} and \texttt{motion} targets
are similarly effective, which is somewhat out of expectation.
{\bf (2)} When they are combined, a clear performance gain can be obtained, suggesting their good complementing effect.
Further, as shown in Table~\ref{tab:targets}, their combining weight is only slightly sensitive to the performance.
{\bf (3)} Regarding the motion temporal granularity,
the results in Table~\ref{tab:motion range} verify that
the most fine-grained motion (\ie, obtained by contrasting the directly adjacent frames) is most useful, as expected.
A plausible reason is that predicting large-time-gap motion is over challenging due to more complex dynamic variations over time.

\paragraph{Decoder design}
Another key component in MAE is the design of decoder. 
To test this, we vary the capacity of our decoder in terms of width and depth.
As shown in Table~\ref{tab:dim} and Table~\ref{tab:blocks}, it is important that both the width (\eg, 384 dimensions) and depth (\eg, 4 blocks) need to be sufficiently large but not too many.

\paragraph{Reconstruction loss function}
Lastly, we evaluate the effect of reconstruction loss function.
We consider three options: $L_1$, \texttt{Smooth $L_1$}, and Mean Squared Error (\texttt{MSE}).
As shown in Table~\ref{tab:loss}, 
\texttt{MSE} yields the best result among these.

\subsection{Model Behavior Visualization }
\begin{figure}[t]
\vspace{-2em}
    \centering
    \subfloat[Something-something V2]{
    \includegraphics[width=0.48\textwidth]{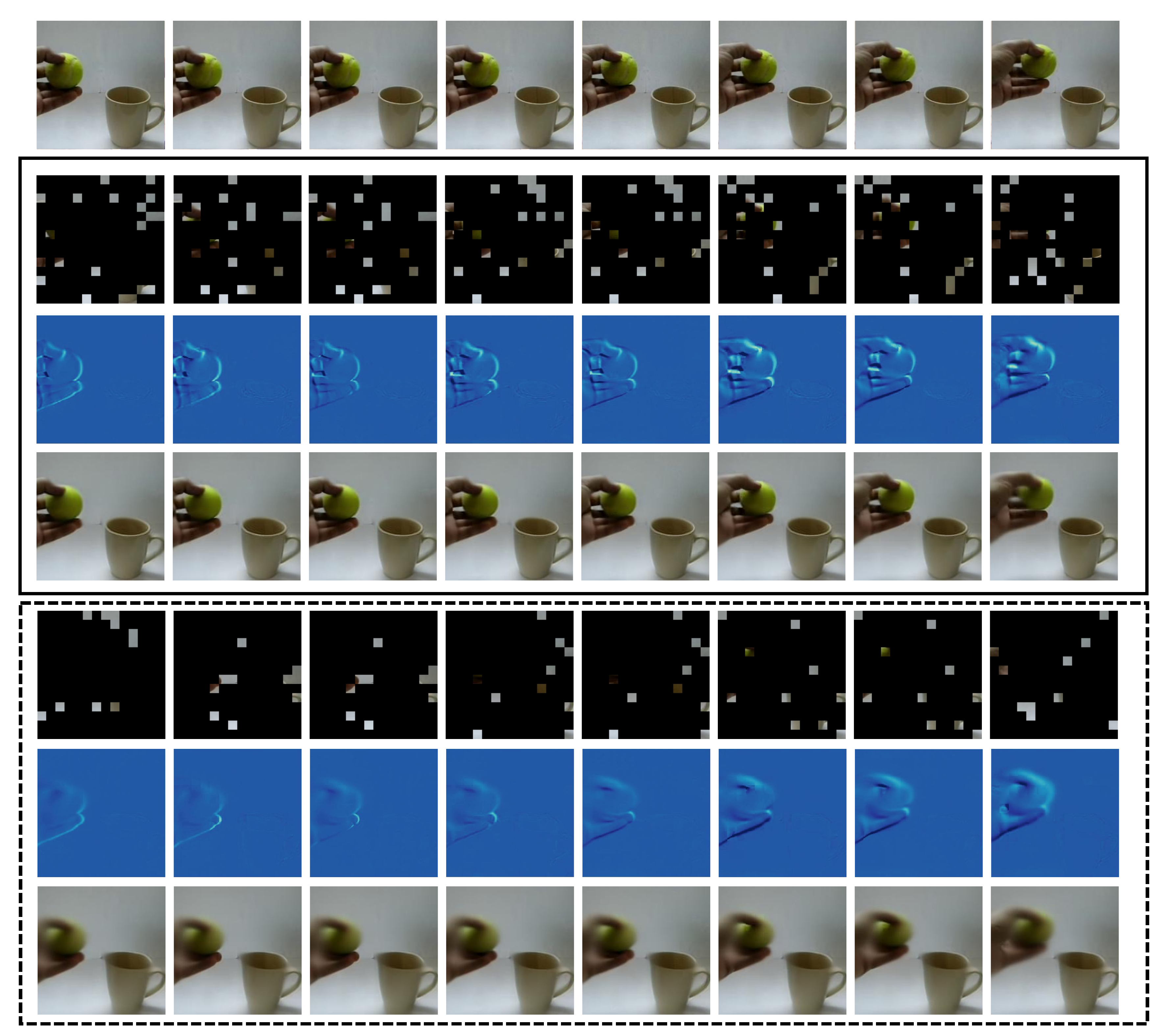}
     \includegraphics[width=0.48\textwidth]{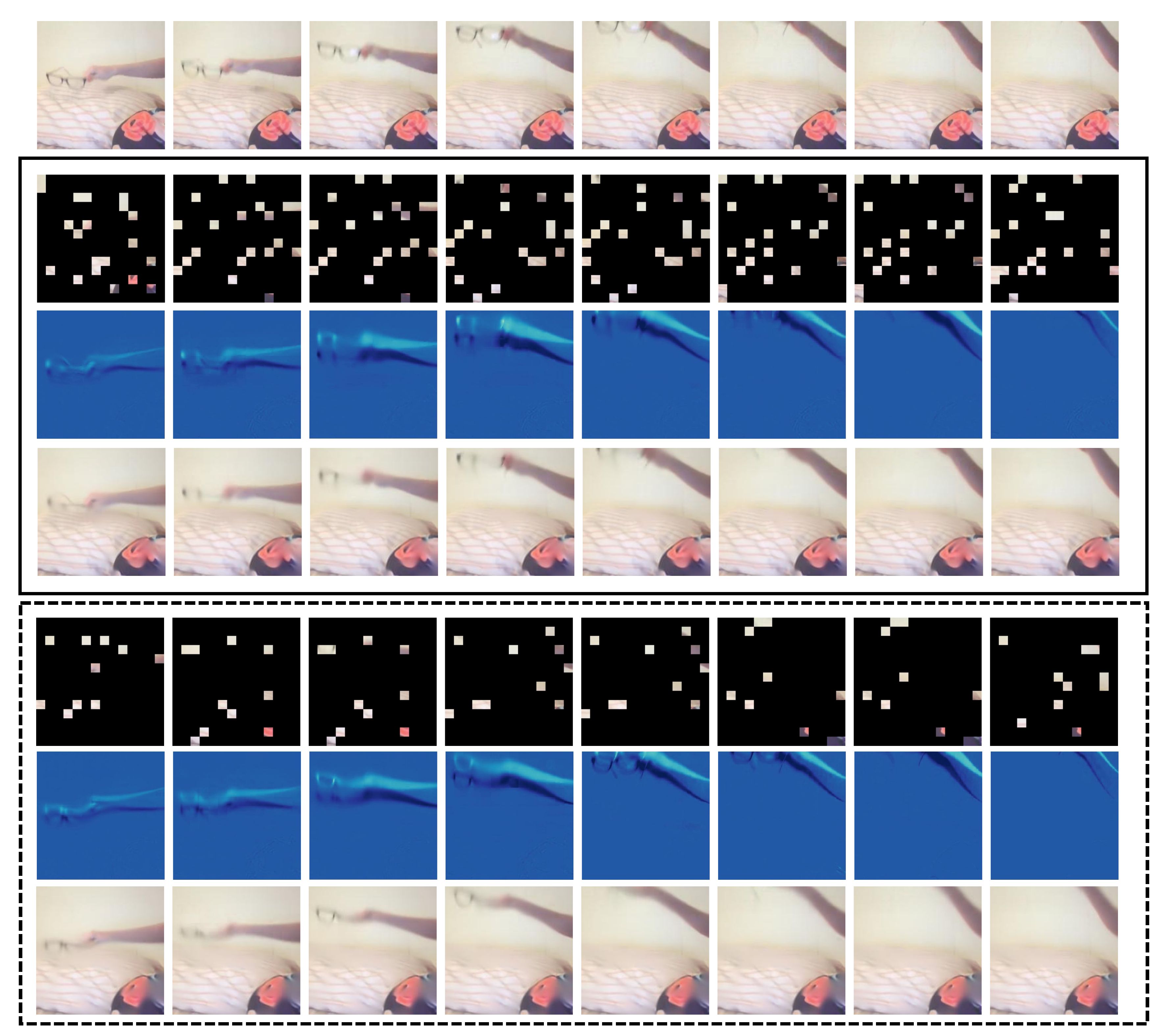}
    }
    \vspace{.2em}
    \centering
    \subfloat[UCF101]{
    \includegraphics[width=0.48\textwidth]{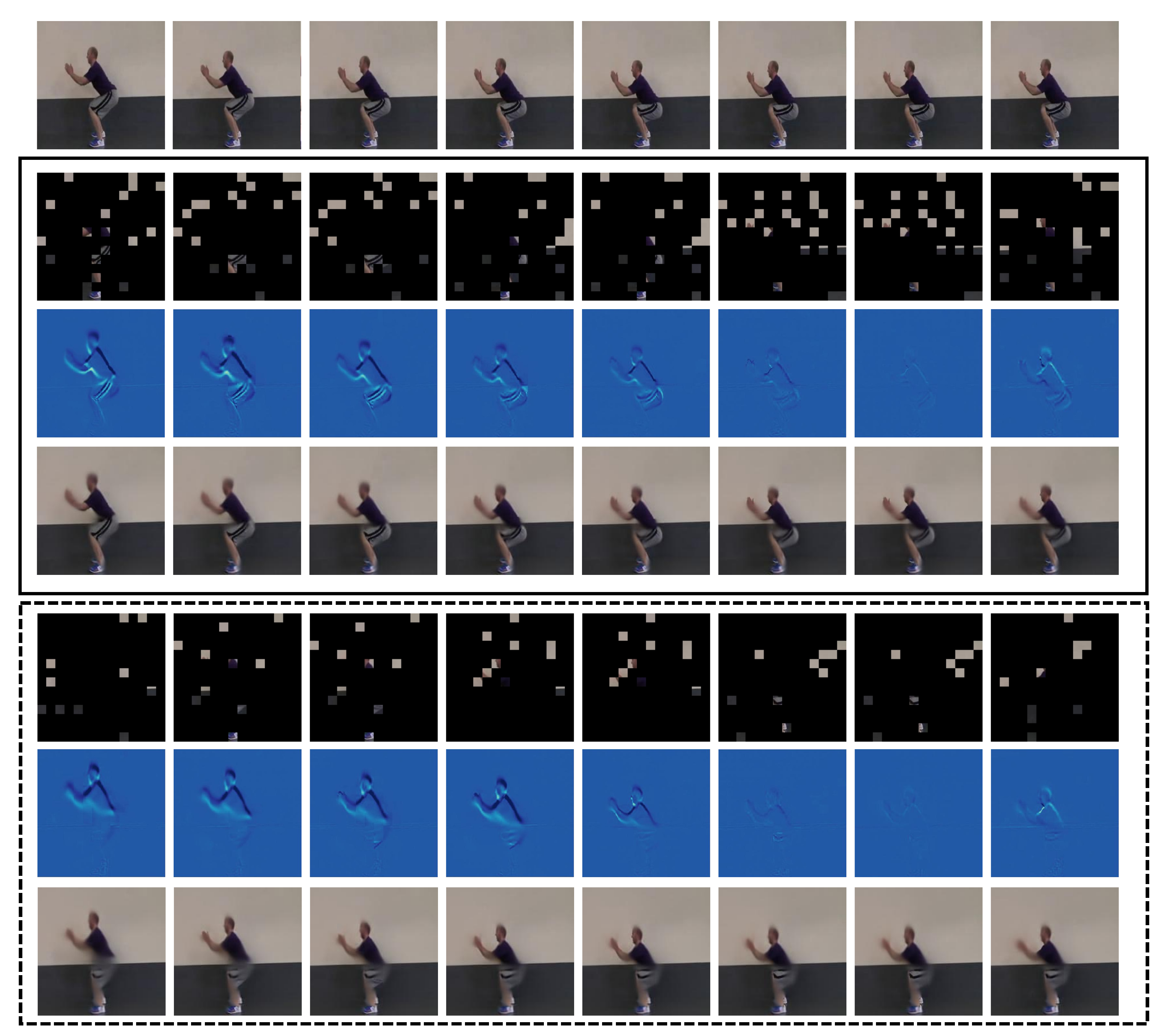}
    }
    \vspace{.2em}
    \centering
    \subfloat[Kinetics-400]{
    \includegraphics[width=0.48\textwidth]{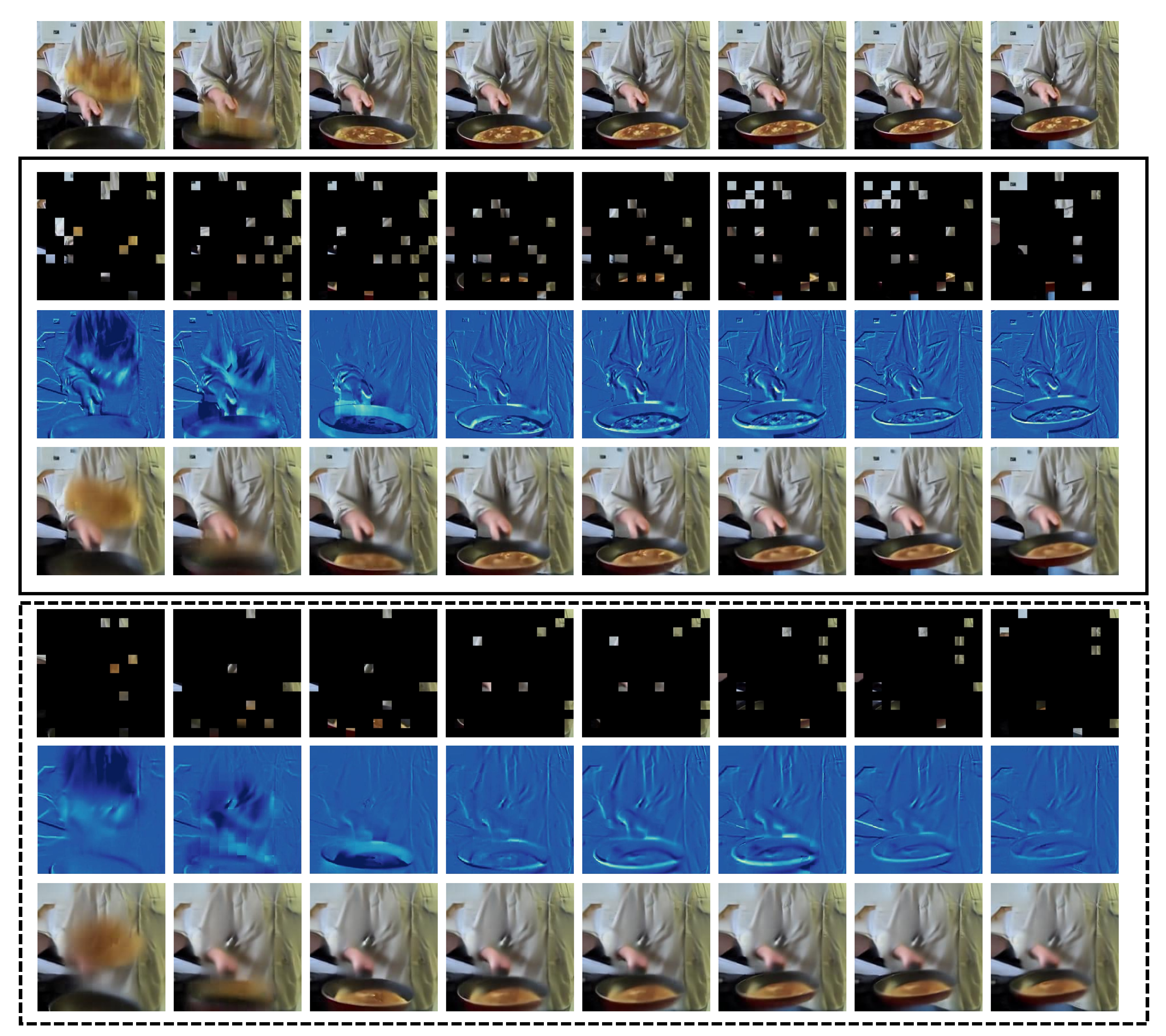}
    }
\caption{Example reconstruction of our \texttt{MotionMAE} on the validation set of (a) Something-Something v2, (b) UCF101, and (c) Kinetics-400.
For each dataset, the {\em first} row shows the original video clip, and two {\em masking ratios} are visualized: \textbf{90\%} (solid box) and \textbf{95\%} (dashed box).
Best viewed with zooming-in.
}
\label{fig:main_visualization}
\end{figure}

To examine the model pretraining behavior, we
visualize the reconstruction output. 
As shown in Figure ~\ref{fig:main_visualization},
under 90\% masking, our \texttt{MotionMAE} model can well generalize the reconstruction ability to new video samples for both individual video frames
and dynamic inter-frame motion.
When increasing the masking ratio to 95\%,
the reconstruction results still preserve the most information 
about both appearance and motion although less fine-grained.
This verifies that our model can recover well the masked data 
to support proper self-supervised learning of video representation.


\section{Conclusion}
In this paper, we have presented motion-aware Masked Autoencoders, namely \texttt{MotionMAE}, for self-supervised video representation learning.
Apart from learning to reconstruct individual masked patches of video clips, our model is designed particularly to concurrently predict the corresponding motion structure information over time. 
Compared to prior art MAEs, our model can better perceive both static appearance and dynamic motion spontaneously, yielding superior spatiotemporal representation learning.
Empirical results on standard action recognition benchmarks demonstrate that our \texttt{MotionMAE} outperforms significantly both supervised learning baseline and state-of-the-art MAE alternatives, under both domain-specific and domain-generic {\em pretraining}-then-{\em finetuning} settings.
%
The generalization and transferability 
of our pretrained models have been also evaluated on another
challenging motion-centric task, video object segmentation.

\bibliography{iclr2023_conference}
\bibliographystyle{iclr2023_conference}

\newpage
\appendix
\section{Additional Implementation Details}
\begin{table}[h!]\centering
\subfloat[{Pretraining}\label{tab:detail_pt}]{
\tablestyle{3pt}{1.00}
\begin{tabular}{y{85}|x{75}x{75}x{75}}
config  & Something-Something V2 & Kinetics400 & UCF101 \\
\shline
optimizer & \multicolumn{3}{c}{AdamW~\citep{loshchilov2017decoupled}} \\
base learning rate  & \multicolumn{3}{c}{1.5e-4} \\
weight decay & \multicolumn{3}{c}{0.05} \\
optimizer momentum & \multicolumn{3}{c}{$\beta_1, \beta_2{=}0.9, 0.95$~\citep{chen2020generative}} \\
learning rate schedule & \multicolumn{3}{c}{cosine decay~\citep{loshchilov2016sgdr}} \\
warmup epochs & \multicolumn{3}{c}{40} \\
epochs& 2400 & 1600 & 3200 \\
flip augmentation & no & yes & yes \\
batch size & 1024(B)512(L) & 1024(B)512(L) & 128(B)\\
augmentation&  \multicolumn{3}{c}{MultiScaleCrop} \\
patch norm & no & yes & yes \\
masking ratio & 90\% & 90\% & 75\% \\
masking type  & random & tube & random \\

\end{tabular}
}
\hspace{1em}
\subfloat[{Finetuning}\label{tab:detail_ft}]{
\tablestyle{3pt}{1.00}
\begin{tabular}{y{85}|x{75}x{75}x{75}}
config  & Something-Something V2 & Kinetics400 & UCF101 \\
\shline
optimizer & \multicolumn{3}{c}{AdamW~\citep{loshchilov2017decoupled}} \\
base learning rate  & 5e-4 & 5e-4(B)2e-3(L) &1e-3 \\
weight decay & \multicolumn{3}{c}{0.05} \\
optimizer momentum & \multicolumn{3}{c}{$\beta_1, \beta_2{=}0.9, 0.999$} \\
layer-wise lr decay &  \multicolumn{3}{c}{0.75~\citep{bao2021beit}} \\
learning rate schedule & \multicolumn{3}{c}{cosine decay~\citep{loshchilov2016sgdr}} \\
warmup epochs & \multicolumn{3}{c}{5} \\
repeated sampling & 2 & 2 & 1 \\
epochs& 30(B)20(B) & 75(B)35(L)& 100(B) \\
flip augmentation & no & yes & yes \\
batch size & 512(B)256(L) & 384(B)64(L) & 256(B)\\
RandAug &  \multicolumn{3}{c}{(9, 0.5)~\citep{cubuk2020randaugment}} \\
label smoothing&  \multicolumn{3}{c}{0.1~\citep{asano2020labelling}} \\
mixup &  \multicolumn{3}{c}{0.8~\citep{zhang2017mixup}} \\
cutmix&  \multicolumn{3}{c}{1.0~\citep{yun2019cutmix}} \\
drop path &  \multicolumn{3}{c}{0.1(B),0.2(L)} \\

\end{tabular}}
\caption{Settings for model pretraining and finetuning. {\texttt{Note:} \textit{lr} = \textit{base\_lr}$\times$batchsize / 256 per the linear \textit{lr} scaling rule.}}
\end{table}

We pretrain \texttt{MotionMAE} on Something-Something V2, Kinetics-400 and UCF101 using the hyper-parameters as summarized in Table~\ref{tab:detail_pt}. 
The dataset specific hyper-parameters are given in the individual columns, with the others shared across datasets. 
These settings apply to ViT-B and ViT-L, unless specified otherwise. We use the same hyper-parameters for pretraining.
Table ~\ref{tab:detail_ft} summarizes our fine-tuning settings.

We conduct the experiments with 32 A100 GPUs for both pretraining and finetuning on Something-Something V2 and Kinetics-400 datasets. The experiments on smaller UCF101  are trained on 8 V100 GPUs. 
\section{Additional Experimental Results}
\begin{figure}[t]
    \centering
    \subfloat[Performance on Something-something V2]{
    \includegraphics[width=0.48\textwidth]{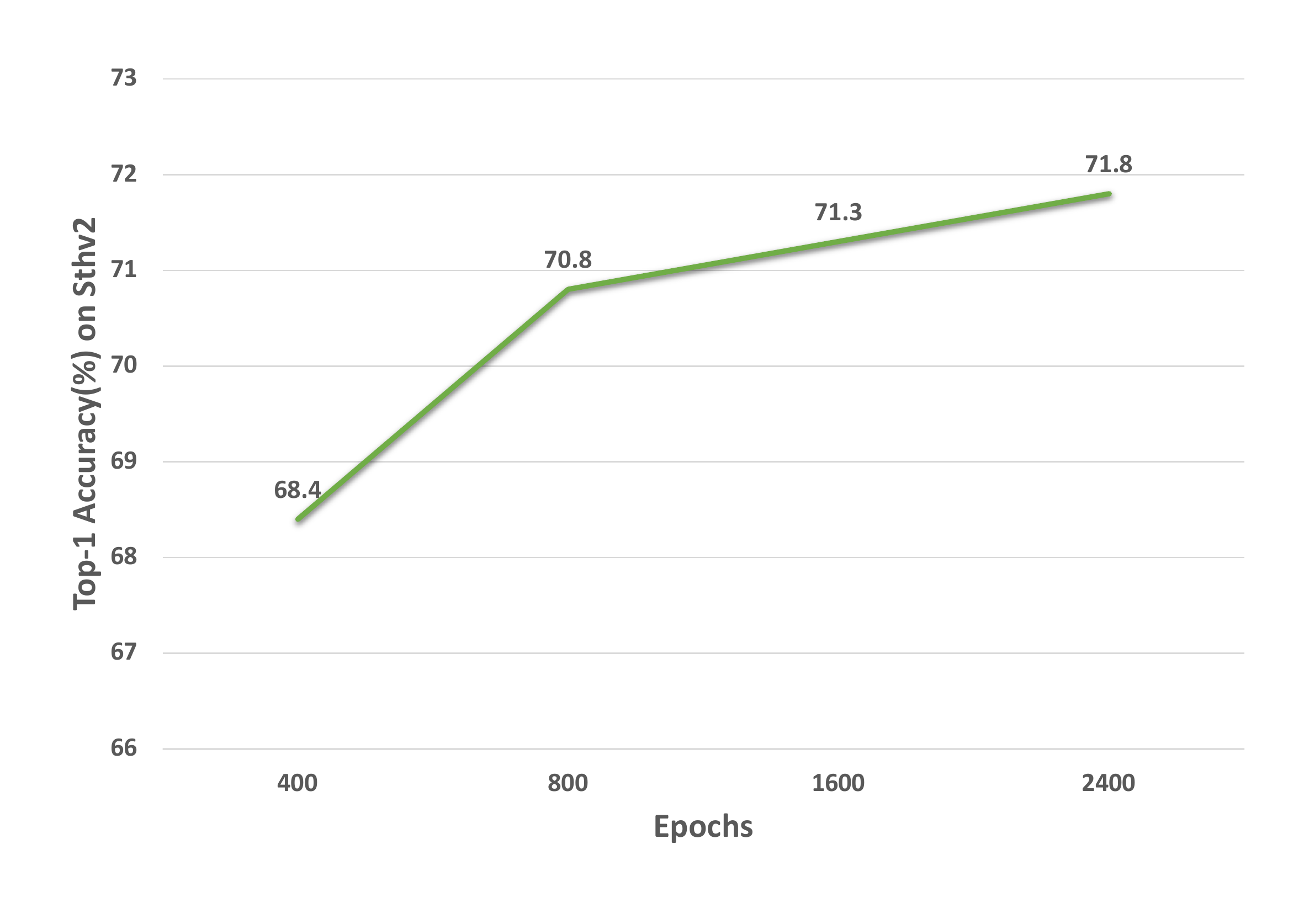}
    }
    \vspace{.2em}
    \centering
    \subfloat[Performance on UCF101]{
    \includegraphics[width=0.48\textwidth]{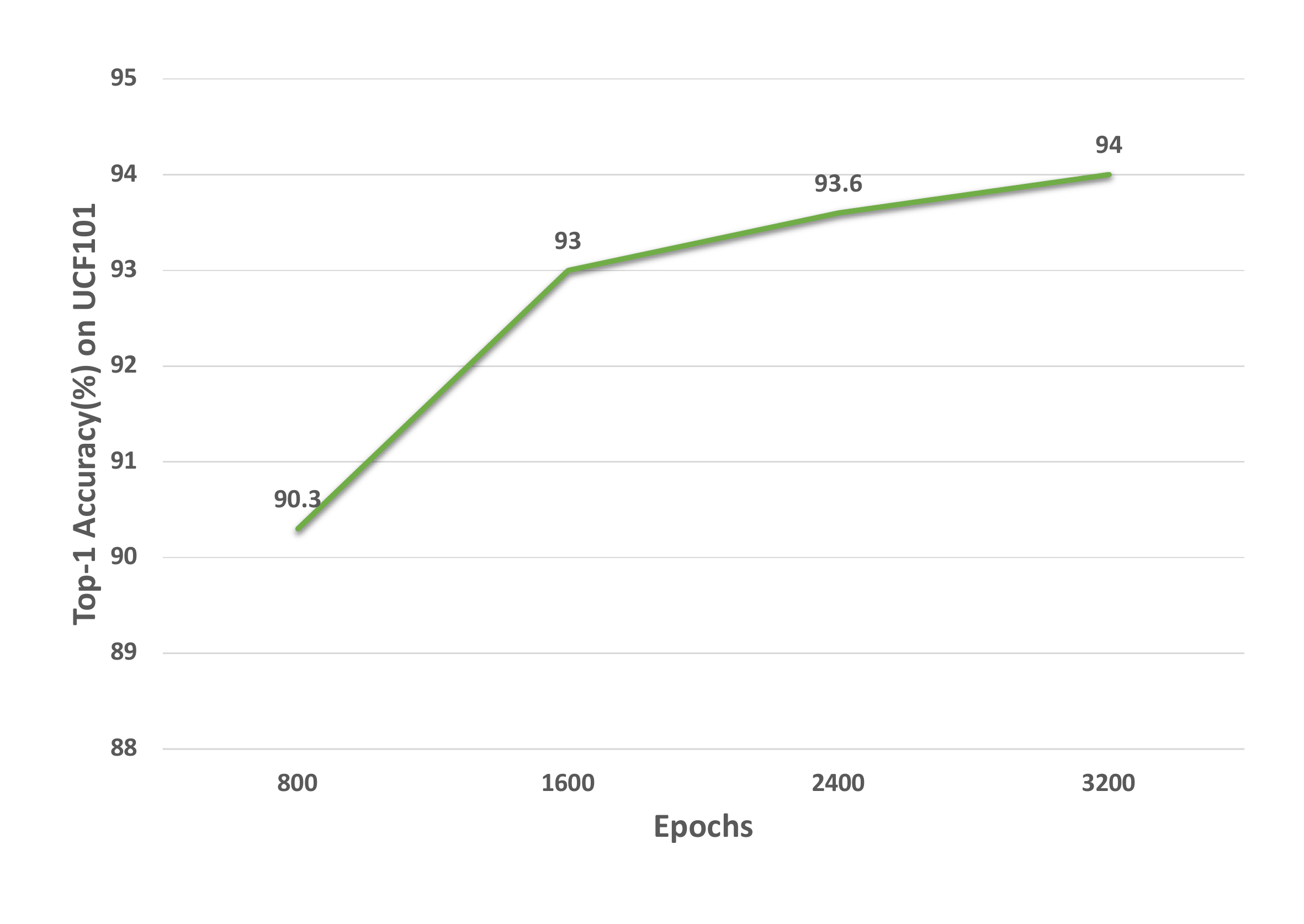}
    }
\caption{
The effect of training schedules on (a) Something-Something V2 and (b) UCF101.
}
\label{fig:training_schedules}
\end{figure} 


\paragraph{Training schedule}
In Figure~\ref{fig:training_schedules} we show the effect of training schedule
on Something-Something (large) and UCF101 (small). 
It is clear that longer training schedule brings slightly gains on both datasets.
More training might lead to further improvement
at extra computation cost.
This indicates high potential of our model.


\paragraph{Local motion extraction: temporal difference vs. optical flow}
\begin{table}[!t]
	\centering
	\begin{minipage}{0.45\textwidth}
	    \begin{center}
		\makeatletter\def\@captype{table}\makeatother
        \begin{tabular}{l | c}
        Target &  top-1 (\%)  \\
        \shline
        Frame  & 93.1  \\
        Frame + OF & 93.5 \\
        Frame + TD & \textbf{94.0}\\
        \end{tabular}
        \vspace{1em}
		\caption{Motion estimation: temporal difference (TD) and optical flow (OF) on UCF101.} 
		\label{tab:diffvsflow}
		\end{center}
	\end{minipage}
	\hspace{1em}
	\quad
	\begin{minipage}{0.45\textwidth}
		\begin{center}
		\makeatletter\def\@captype{table}\makeatother
        \begin{tabular}{c | c}
        Masking strategy &  top-1 (\%)  \\
        \shline
        time-only & 74.6\\
        random  & 78.8 \\
        space-only & \textbf{79.0} \\
        \end{tabular}
        \vspace{1em}
        \caption{Effect of different masking strategies on Kinetics-400.}

		\label{tab:mask_type}
		\end{center}

	\end{minipage}
\end{table}
Except temporal difference (TD) as we have exploited,
optical flow is another popular method for local motion estimation.
It is interesting to see which one is better and why. 
We conduct this test on UCF101.
Specifically, we pre-extract per-frame optical flow (OF)
to replace the TD during model pre-training, whilst keeping the others the same.
As shown in Table~\ref{tab:diffvsflow}, 
we can see that OF is relatively inferior to TD, 
whilst still improve over using frames only.
%
%
This is seemingly out of expectation since OF is often considered to be more accurate motion estimation at much more computational cost.
We consider that the underlying reason is that
the frames used to compute OF is not fully available 
in the input.
Suppose the $t$-th frame is sampled to a video clip, the $t+1$-th frame which is used to compute OF, is however excluded by our sampling strategy (see \texttt{data pre-processing} in Section \ref{sec:impl}).
This discrepancy may cause the lower performance by OF.
In contrast, our TD has no such problem whilst being more efficient.

\paragraph{Mask sampling strategy}
We evaluate three masking strategies (random, time-only, space-only) on Kinetics400.
As shown in Table~\ref{tab:mask_type}, we find that random and space-only are similarly performing,
whilst time-only is the worst.
This is not surprising since reconstructing the whole frames could be over challenging.

\paragraph{Masking ratio}
We evaluate the effect of mask ratio on 
Something-something V2 (large) and UCF101 (small).
We adopt the data-specific pretraining setting. 
As shown in Table~\ref{tab:ratio},
this setting is considerably dataset (\eg, size) specific.
This is not observed in the previous works.

\begin{table}[!t]
	\centering
	\begin{minipage}{0.45\textwidth}
	    \begin{center}
		\makeatletter\def\@captype{table}\makeatother
        \begin{tabular}{c | c  c}
        Ratio &  UCF & Something-Something V2  \\
        \shline
        75\% & \textbf{94.3} & 66.0 \\
        90\% & 92.7 & \textbf{68.4} \\
        \end{tabular}
        \vspace{1em}
		\caption{Effect of masking ratio.} \label{tab:ratio}
		\end{center}
	\end{minipage}
	\hspace{1em}
	\quad
	\begin{minipage}{0.45\textwidth}
	    \begin{center}
		\makeatletter\def\@captype{table}\makeatother
        \begin{tabular}{c | c}
        Arch &  top-1 (\%)  \\
        \shline
        Att. share & 68.0\\
        Divide  & \textbf{68.4}  \\
        \end{tabular}
        \vspace{1em}
		\caption{Effect of decoder architecture on Something-Something V2.} 
		\label{tab:arc}
		\end{center}
	\end{minipage}
	
\end{table}

\paragraph{Decoder architecture}
We evaluate the effect of decoder architecture design.
By default, we adopt a {\em parallel} structure
with two independent networks: a space head to predict masked frame patches, and a time head to predict the corresponding local motion structure.
An alternative is to {\em share} a network except the last prediction layer.
As shown in Table~\ref{tab:arc}, 
network sharing is less effective, 
confirming our design choice.

\section{Additional Visualization Examples}
\begin{figure}[t]
    \centering
    \includegraphics[width=0.48\textwidth]{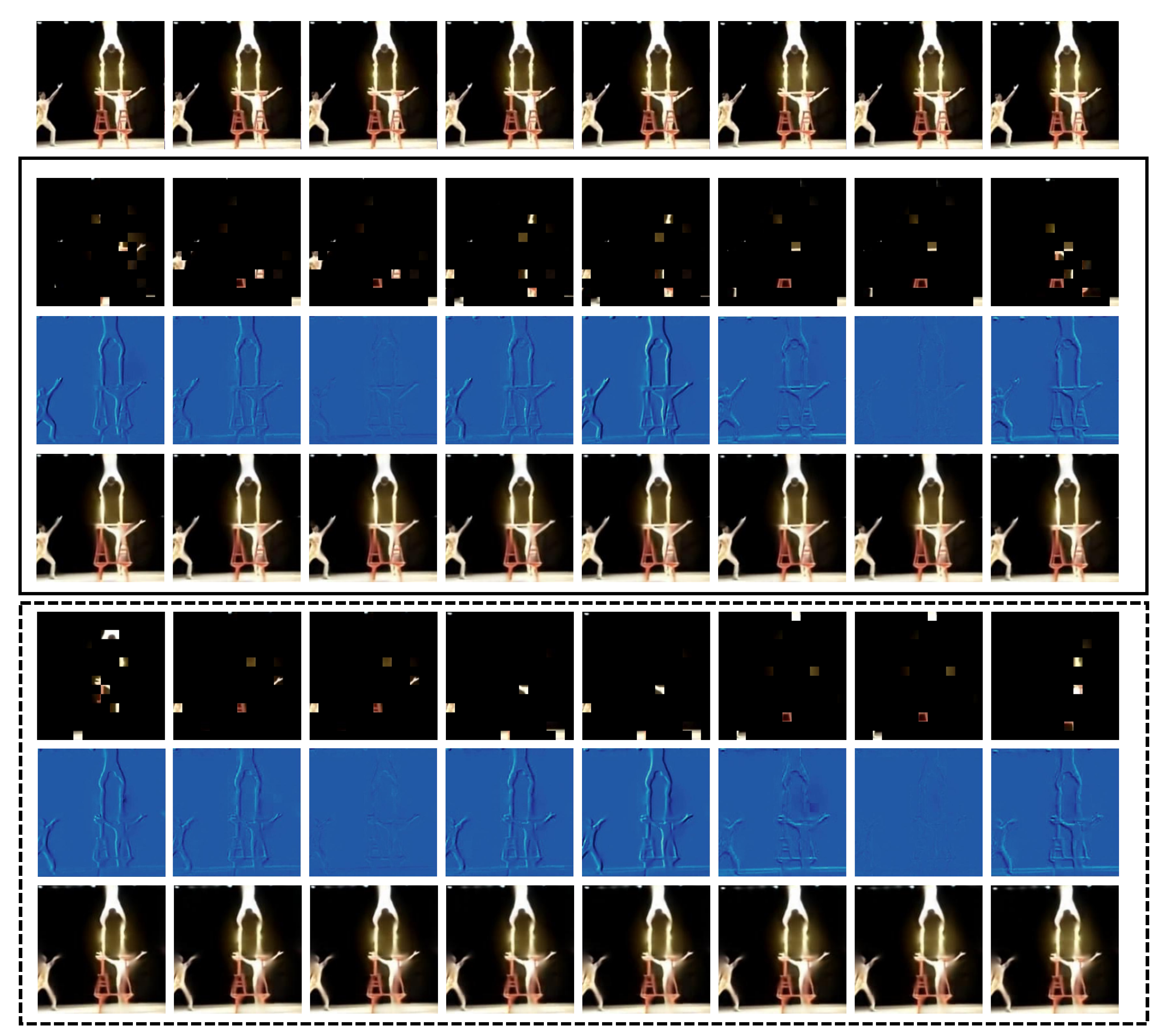}
    \includegraphics[width=0.48\textwidth]{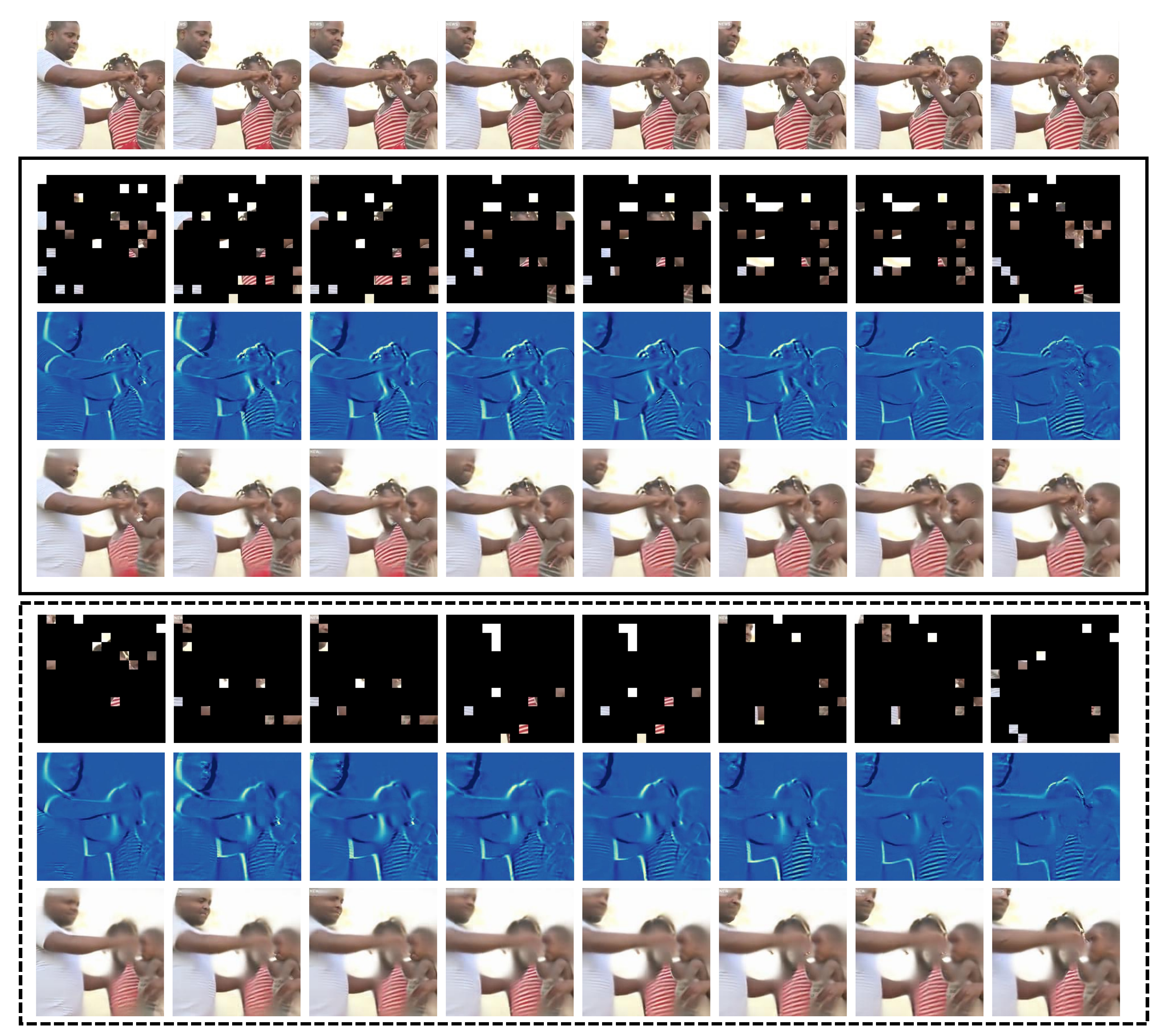}
\caption{
More visualizations on Kinetics-400.
The {\em first} row shows the original video clip, and two {\em masking ratios} are visualized: \textbf{90\%} (solid box) and \textbf{95\%} (dashed box).
Best viewed with zooming-in.
}
\label{fig:figvis_supp_2}
\end{figure}

\begin{figure}[]
    \subfloat[Something-something V2]{
    \begin{minipage}[b]{1 \linewidth}
		 \includegraphics[width=0.5\linewidth]{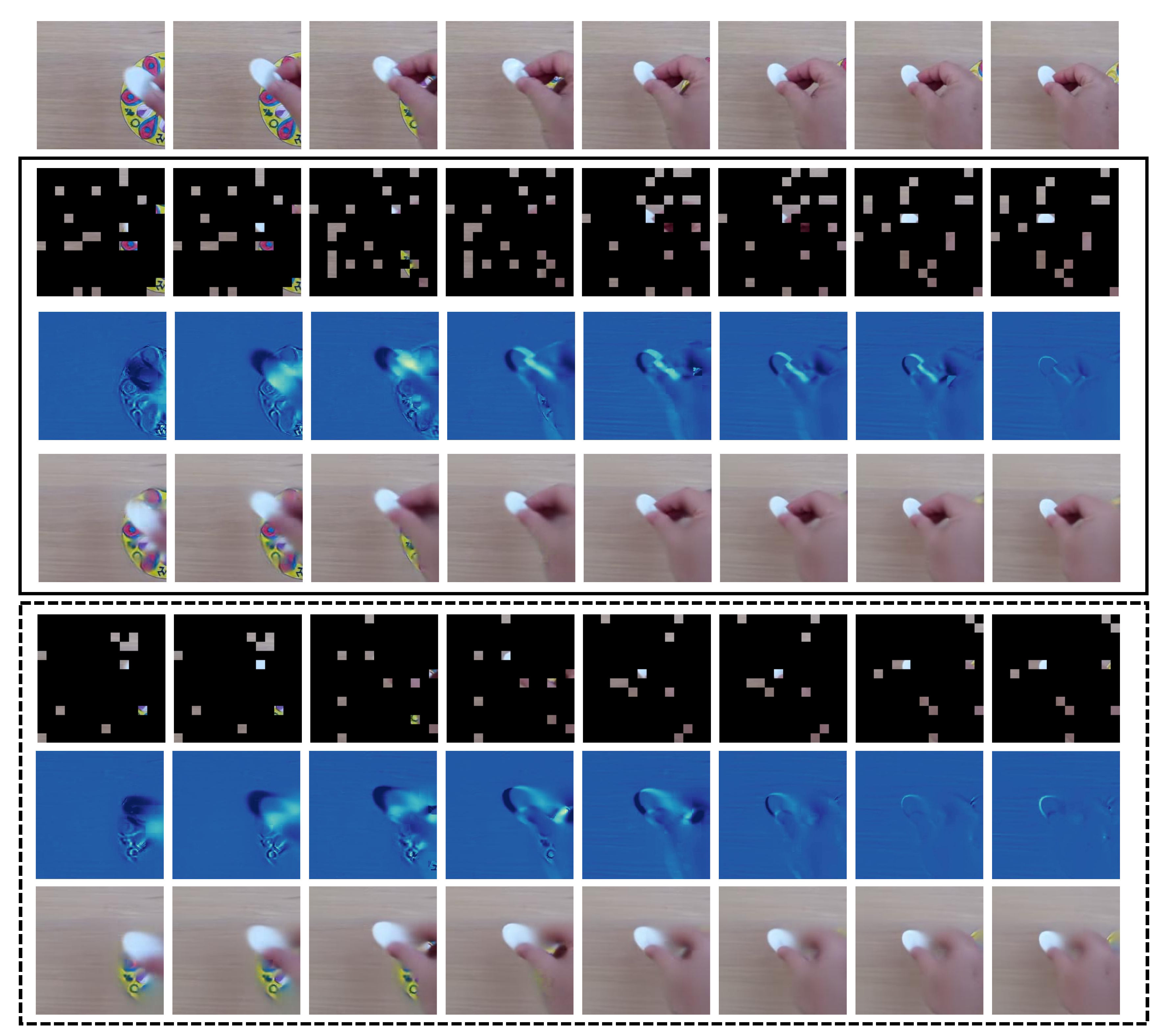}
         \includegraphics[width=0.5\linewidth]{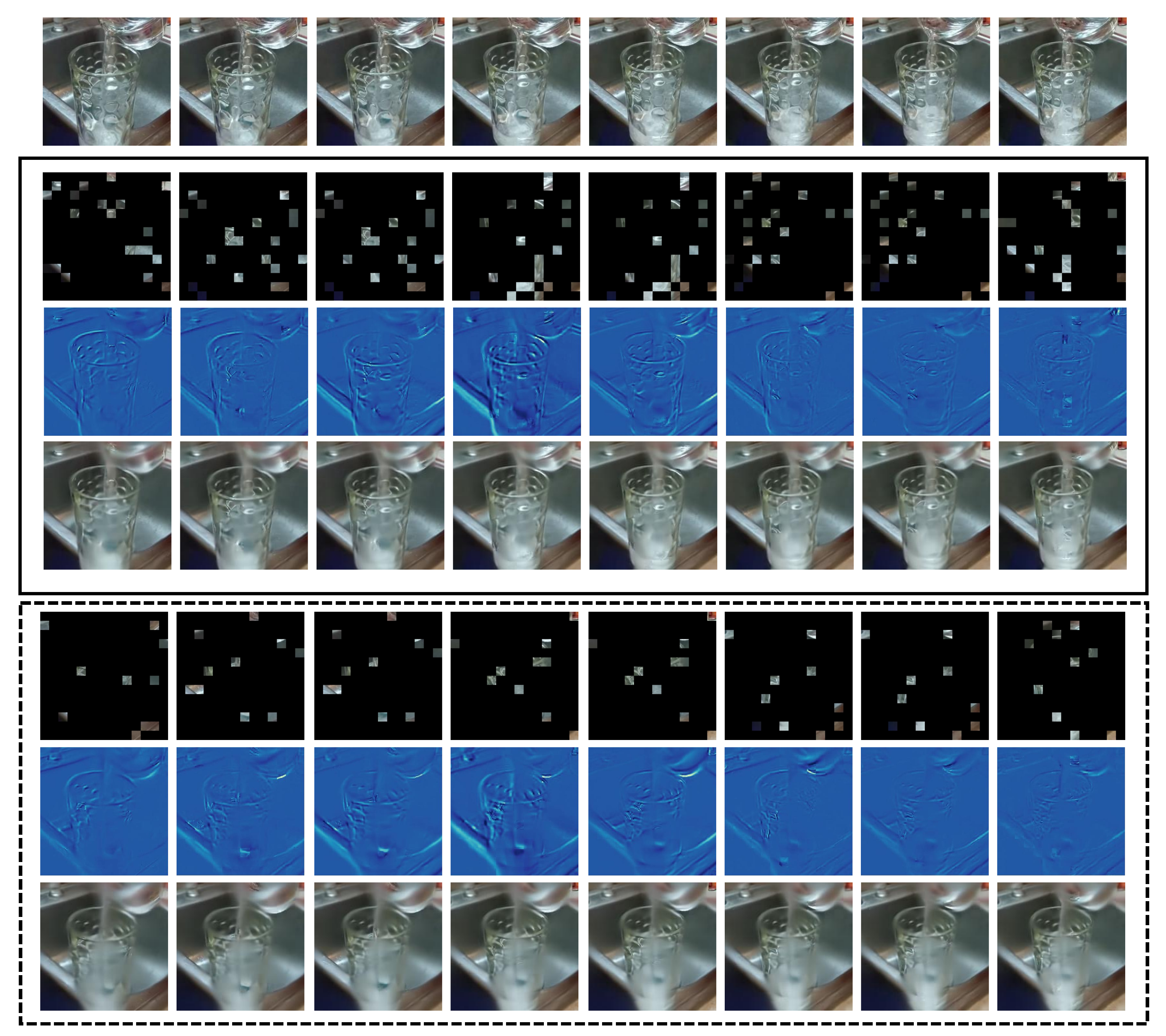}
         \includegraphics[width=0.5\linewidth]{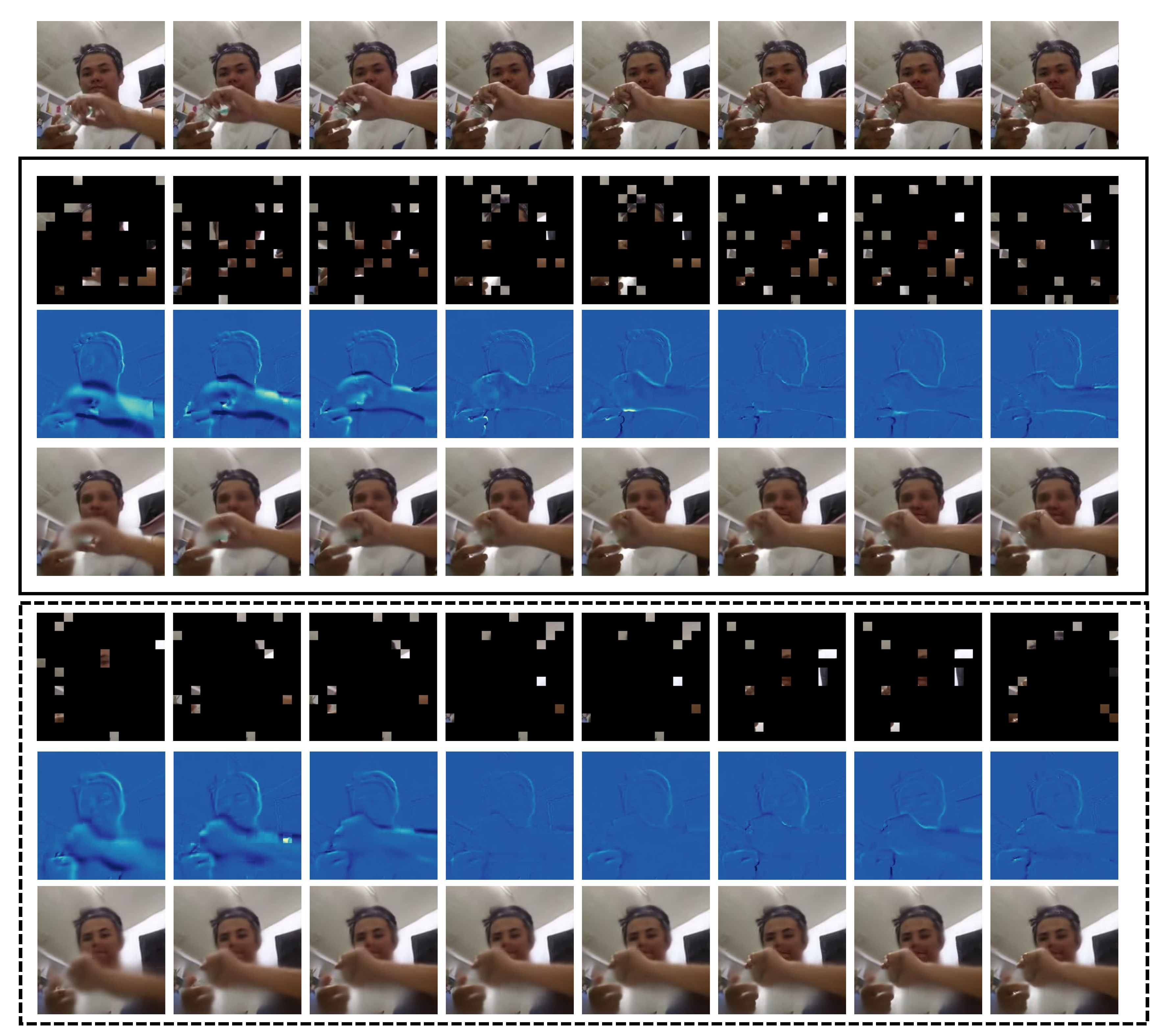}
          \includegraphics[width=0.5\linewidth]{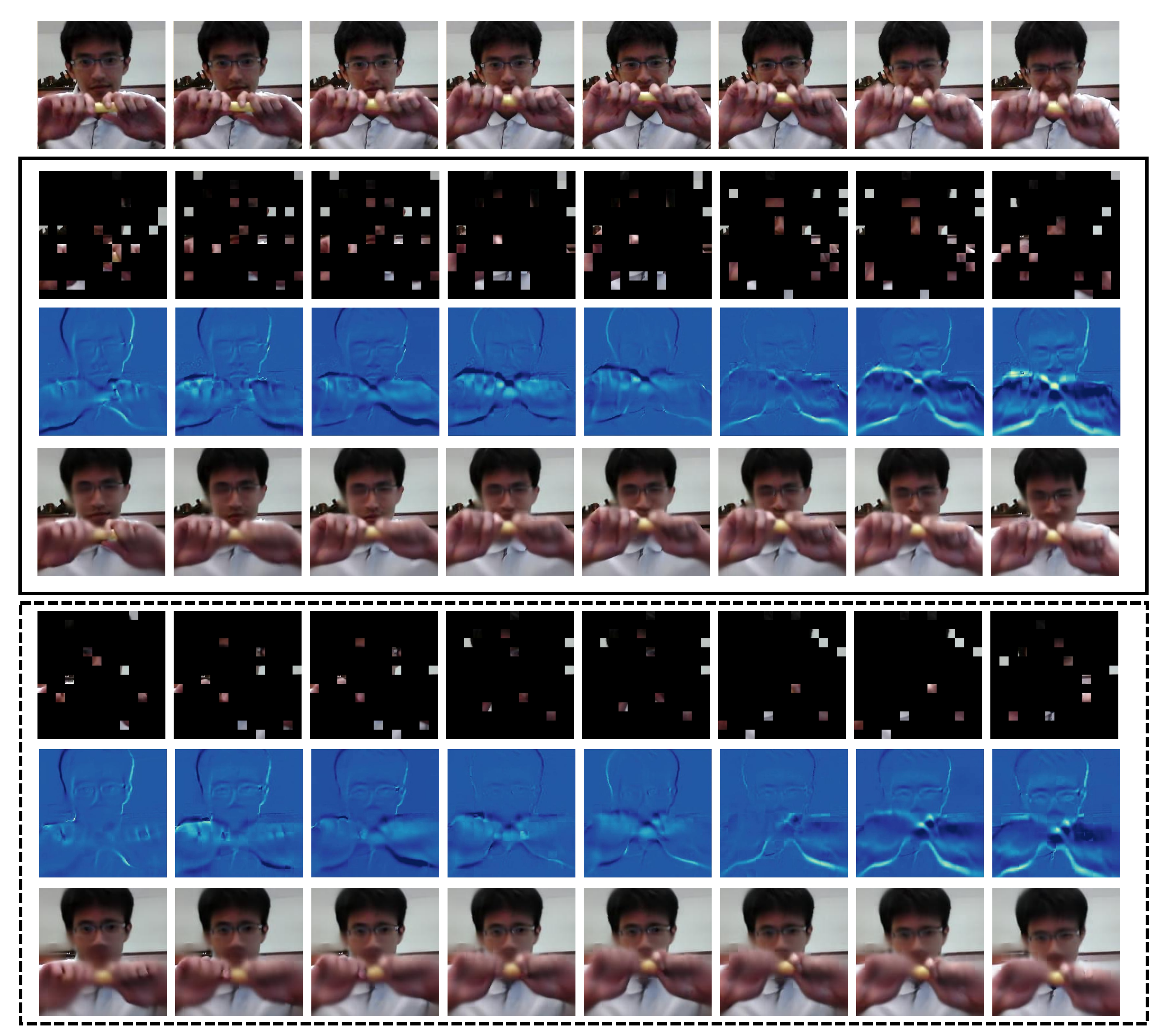}
	\end{minipage}
	}
	\quad
	\centering
    \subfloat[UCF101]{
    \begin{minipage}[b]{1 \linewidth}
		 \includegraphics[width=0.5\linewidth]{./supp_vis/sthv2_vs_supp_0.pdf}
         \includegraphics[width=0.5\linewidth]{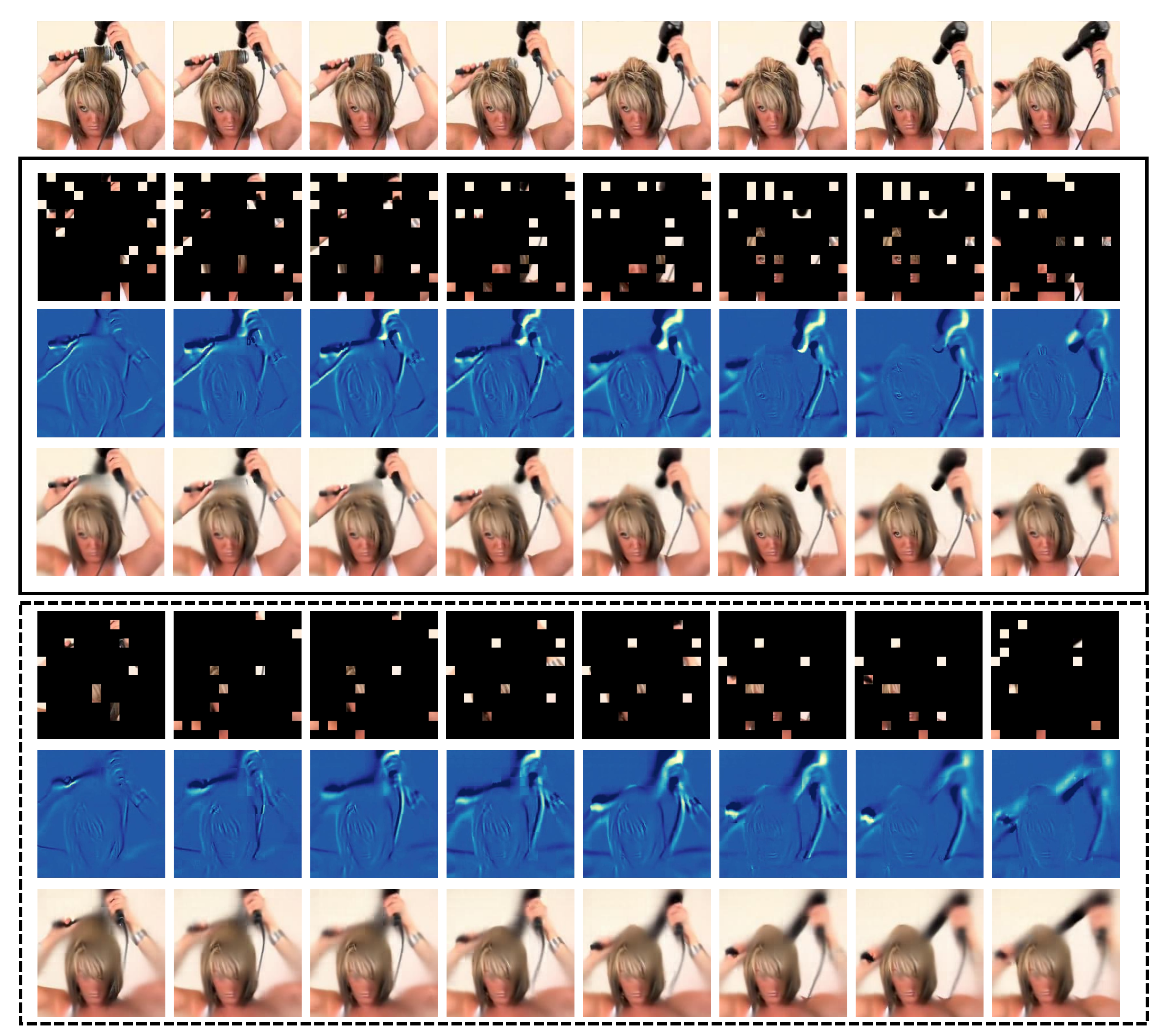}

	\end{minipage}
    }
   
\caption{More visualizations on Something-something V2 and UCF101.
For each dataset, the {\em first} row shows the original video clip, and two {\em masking ratios} are visualized: \textbf{90\%} (solid box) and \textbf{95\%} (dashed box).
Best viewed with zooming-in.
}
\label{fig:figvis_supp_1}
\end{figure}

Figure~\ref{fig:figvis_supp_1} and ~\ref{fig:figvis_supp_2} provide more examples of reconstruction on Something-something V2, UCF101 and Kinetics-400.
\newpage
\end{document}